\pdfoutput=1

\documentclass[11pt]{article}

\usepackage[review]{EMNLP2023}

\usepackage{times}
\usepackage{latexsym}

\usepackage[T1]{fontenc}

\usepackage[utf8]{inputenc}

\usepackage{microtype}

\usepackage{inconsolata}
\usepackage{graphicx}
\usepackage{enumitem}
\usepackage{subcaption} 
\usepackage{pgfplots} 
\usepackage{multirow}

\title{FEDERATED LEARNING BASED MULTILINGUAL EMOJI PREDICTION IN CLEAN AND ATTACK SCENARIOS}

\author{First Author \\
  Affiliation / Address line 1 \\
  Affiliation / Address line 2 \\
  Affiliation / Address line 3 \\
  \texttt{email@domain} \\\And
  Second Author \\
  Affiliation / Address line 1 \\
  Affiliation / Address line 2 \\
  Affiliation / Address line 3 \\
  \texttt{email@domain} \\}

\begin{document}
\nolinenumbers
\maketitle
\begin{abstract}
Federated learning is a growing field in the machine learning community due to its decentralized and private design. Model training in federated learning is distributed over multiple clients giving access to lots of client data while maintaining privacy. Then, a server aggregates the training done on these multiple clients without access to their data, which could be emojis widely used in any social media service and instant messaging platforms to express users' sentiments. This paper proposes federated learning-based multilingual emoji prediction in both clean and attack scenarios. Emoji prediction data have been crawled from both Twitter and SemEval emoji datasets. This data is used to train and evaluate different transformer model sizes including a sparsely activated transformer with either the assumption of clean data in all clients or poisoned data via label flipping attack in some clients. Experimental results on these models show that federated learning in either clean or attacked scenarios performs similarly to centralized training in multilingual emoji prediction on seen and unseen languages under different data sources and distributions. Our trained transformers perform better than other techniques on the SemEval emoji dataset in addition to the privacy as well as distributed benefits of federated learning. 
\end{abstract}

\section{Introduction}

Federated learning FL is an emerging field in machine learning first introduced by Google in 2017 \citep{mcmahan2017communication}. FL is a distributed learning framework, where multiple distributed clients collaborate in training machine learning models under the orchestration of a central server. This server aggregates the trained models into a final global model. FL is a major shift from centralized
machine learning to a distributed manner that
uses many distributed computing resources such as devices and data. Policies such as General
Data Protection Regulation (GDPR) \citep{regulation2018general} imposes data privacy rules among different organizations. Thus, FL is essential to enhance data privacy by keeping the raw data on the local device while taking into account that some clients may have a Non-IID distribution of data or poisoned data limiting federated learning accuracy.

Federated learning is used in many natural language processing applications \citep{li2020federated, singh2022federated} such as adapting to pedestrian behavior in data generated by distributed sensors and devices, enabling real-time analysis and decision-making. In this paper, we focus on the task of emoji prediction because emojis enhance communication quality among users and associated text data is private. For example, the use of emojis in tweets has been steadily increasing over the years, with 35\% of tweets in 2019 \citep{mcshane2021emoji} containing at least one emoji, compared to 9.9\% in 2012. This trend has led to a 25\% increase in engagement for tweets that feature at least one emoji \citep{mcshane2021emoji}, as compared to those without any emojis. Moreover, the use of emojis in combination with brand names has increased by 49\% since 2015, \citep{agnew2017emoji}. 


While traditional next-word prediction models are limited by the relatively narrow range of options available, emoji prediction presents unique challenges due to the diverse and context-dependent nature of emoji usage. For example, when predicting the next word after 'you' in the sentence 'Thank you so much', the options are relatively limited and predictable, such as 'guys' or 'so'. However, for the next emoji prediction after 'you', the options are more diverse and context-dependent, ranging from different colored hearts to facial expressions. 


Reviewing the literature on emoji prediction, we can 
find that several papers indeed proposed methods in either centralized or federated settings \citep{semeval2018task2, weller2022pretrained, ramaswamy2019federated, gandhi2022federated, lee2022multiemo, tomihira2020multilingual, peng2021seq2emoji, barbieri2018multi, yang2018applied, barbieri2020tweeteval, venkit2021asourceful, edwards2020go, camacho2022tweetnlp, barbieri2022xlm, loureiro2022timelms, caldarola2022improving}. For example, SemEval shared task in \citep{semeval2018task2} is dedicated to emoji prediction where multiple methods were developed in a centralized setting with a maximum F1 score of 35.99\%. In this shared task, methods did not consider the widely used transformer architecture, multilinguality, and federation. In addition, Google successfully implemented a federated learning solution for Gboard's emoji prediction using the LSTM architecture in \citep{ramaswamy2019federated}. However, the results in this paper were only shown in English with 100 emoji classes while assuming that all clients have clean data, i.e clean scenarios. Additionally, the findings presented in \citep{weller2022pretrained} indicate that multilingual federated learning can be achieved without significant performance degradation compared to centralized learning. However, the study did not evaluate the performance of their federated learning models in the case of unseen languages or poisoned data from some clients (i.e., attack scenarios). Notably, the study did not include sparsely activated MoE transformers \citep{kim2021scalable}, which showed success in NLP tasks. Although previous work by \citep{gandhi2022federated} achieved good results in predicting emojis in Hindi tweets using both centralized and federated learning approaches, our research aims to expand upon this approach by applying it to a multilingual context. It is worth noting that their study did not explore the effects of sparsely activated MoE transformers or evaluate the performance of their federated models on unseen languages or under poisoned data attack scenarios.

In a parallel research line, there were studies that investigated FL in an attack scenario, and proposed techniques to defend against label-flipping attacks and backdoor attacks \citep{lyu2020threats, rodriguez2022dynamic, blanchard2017machine, wang2020attack, fung2020limitations, jebreel2022defending, manoel2022federated, ma2020safeguarding}. However, their experiments were not carried out on the task of multilingual emoji prediction with multiple classes, and the results were not compared to the centralized setting. 

This paper proposes federated learning-based multilingual emoji prediction in both clean and attack scenarios\footnote{Demo and Source code of this paper on GitHub \href{https://github.com/kareemgamalmahmoud/FEDERATED-LEARNING-BASED-MULTILINGUAL-EMOJI-PREDICTION-IN-CLEAN-AND-ATTACK-SCENARIOS}{( FEDERATED-LEARNING-BASED-MULTILINGUAL )}} Our two million training and testing examples are acquired from both Twitter and the standard SemEval emoji dataset \href{https://competitions.codalab.org/competitions/17344}{(SemEval Data)}. For multilingual emoji prediction, we train publicly available pre-trained models of different sizes of dense and sparsely activated transformers, namely, Multilingual-MiniLM (M-MiniLM) \citep{wang2020minilm}, Twitter-twihin-Bert-base (Bert-Base) \citep{zhang2022twhin}, Twitter-XLM-Roberta (XLM-R) \citep{barbieri2022xlm}, and switch-MoE with 8 experts \citep{fedus2021switch}. To simulate attack scenarios in FL, we apply the label-flipping data poisoning attack to some clients and utilize different FL aggregation schemes to reverse this attack. Our experimental results led to the following findings:

\begin{itemize}[noitemsep,nolistsep]
    \item In either centralized or FL experiments, we achieved emoji prediction accuracy better than the teams reported in the SemEval emoji prediction shared task \citep{semeval2018task2}.
    
    \item FL training on emoji data achieves similar accuracy performance to traditional centralized setup. This FL accuracy performance is confirmed in seen or unseen languages with IID and Non-IID data distributions in both unilingual and multilingual settings.
    
    \item When some clients' emoji data is attacked via label flipping, FL's K-representative unweighted median (Krum)  aggregation scheme can restore the accuracy of the clean setting. 
\end{itemize}

The rest of the paper is organized as follows: Section \ref{sec:method} describes the data, models, training algorithm, and label-flipping attack for FL. This is followed by explaining experiments carried out in this paper. Last but not least, Section \ref{sec:Conclusion} concludes the paper.

\section{Methodology}
\label{sec:method}

\subsection{Data Acquisition}

\begin{figure} [http]
\setlength{\belowcaptionskip}{-4ex}
      \centering
             \includegraphics[width=1\linewidth]{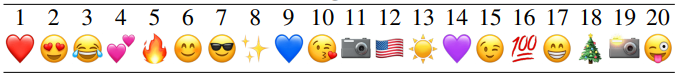}
      \caption{The 20 most frequent emojis.}
      \label{fig:emojis_list}
\end{figure}

We used the Twitter API to crawl over 2 million tweets that contain only one emoji as well as 500k training and test data from SemEval. The Twitter set encompasses 3 languages namely, Spanish, Italian, and French, while SemEval includes English.  This data is filtered to remove stop words, hyperlinks, and duplicate special characters resulting in 1.3 million examples. Given an input sentence, we focus on predicting the 20 most popular emojis, which are the same emojis as the SemEval paper \citep{semeval2018task2} shown in Figure \ref{fig:emojis_list}. 

\subsection{Client Partitioning}
\label{sec:client_part}

Three different training setups were carried out in this paper: traditional centralized training with no FL\footnote{In practical scenarios, publicly available datasets such as Twitter data or datasets similar to the private chat data can be used to train the server model for FL. This is particularly useful in chat problems where users may be hesitant to share their private chat data with the application.}, FL with IID data where each client has a random subset of all data, and FL with Non-IID data where each client includes data from one language. The number of clients for FL experiments is four. We have selected four clients for our FL experiments based on previous experiments and to maintain consistency with the Non-IID setup of our problem. This decision is also influenced by the fact that we initially started our work with four languages. In either FL IID or Non-IID, we carry out an experiment while assuming that all clients have clean data (clean scenario) and another experiment while assuming that 25\% and 50\% of the clients have label-flipped data (attack scenario). For centralized training with no FL, we merge the clients' datasets into one pool and use label flipping to attack the same set of samples as FL to compare the results.

\subsection{Attack Scenario: Label flipping Procedure}

We flipped the first 10 emoji classes in Figure \ref{fig:emojis_list} into the last 10 emoji. Based on this flipping, we created two attack scenarios. The first scenario is to make 1 out of 4 FL clients toxic (i.e, 25\%), while the second scenario is to make 2 out of 4 of FL clients toxic (i.e 50\%). We then apply these datasets to federated training (IID setup, and Non-IID setup) and centralized training as described in Section \ref{sec:client_part}. 

To elaborate on the IID and Non-IID setup, we divided the data into four parts for the IID setup, with each client taking one part. For the 25\% FL clients toxic attack scenario, we flipped the first 10 emoji classes to create a toxic dataset for one of the four clients.

For the Non-IID setup, since we had four languages for training, we divided the data per language and assigned each client one language to work on. In this scenario, we made the English data toxic for the 25\% FL clients attack scenario, while for the 50\% FL clients attack scenario, we made both the English and French languages toxic. This approach allowed us to simulate a more realistic attack scenario, where the toxic data distribution is not uniform across all clients.

\subsection{Federated Learning Methods}

FederatedAveraging (FedAVG) \citep{mcmahan2017communication} is an aggregation scheme used in federated learning, where each client trains a model on its local data using stochastic gradient descent (SGD), and the server aggregates the client weights by taking their average. This process is repeated for a fixed number of rounds, with each round consisting of clients training their models and the server aggregating their weights. On the other hand, Krum \citep{blanchard2017machine} is a robust federated learning algorithm that selects a subset of model updates, excluding those from malicious or incorrect clients, and computes the K-representative unweighted median of the selected updates. The selected update is the one that is closest to the center of the other updates while ignoring the updates that are far away from the center. The parameter K is specified by the user. 

Krum (Krumble) is more computationally demanding than FedAvg (Federated Averaging) because it involves additional steps of distance calculation to select the best model updates.

In Krum, the server selects the best model updates from a subset of the participating devices (clients) based on the distances between the updates. The distances are calculated based on the number of disagreements between the updates and the other updates in the subset. This process requires more computation than FedAvg, where the server simply averages the model updates received from the clients.

Thus, the additional distance calculation step in Krum makes it more computationally demanding than FedAvg. However, Krum may provide better performance in certain scenarios where the participating clients may be potentially malicious or have poor quality updates.

Although, Krum can produce a more precise final model in attack scenarios where FedAVG may not be effective.

\subsection{Models}

Publicly available models with different numbers of model parameters are trained for the task of emoji prediction. In particular, we utilized 4 models from hugging face: (1) M-MiniLM \citep{wang2020minilm} is a 21M parameter transformer model pre-trained on 16 languages and distilled from Bert Base; (2) Bert-Base is a multi-lingual tweet language model \citep{zhang2022twhin} that is trained on 7 billion Tweets from over 100 distinct languages and has 280M parameters; (3) XLM-R \citep{barbieri2022xlm} is trained on ~198M multilingual tweets, has 278M parameters, and pre-trained in more than 30 Languages; (4) switch-MoE is an 8-expert MoE model trained on Masked Language Modeling (MLM) task. The model architecture is similar to the classic T5, but with the Feed Forward layers replaced by the Sparse MLP layers containing "experts" MLP. It has 619M parameters and it's pre-trained in the English-only language. In addition to the publicly available models, we also built an LSTM model from scratch for the emoji prediction task. The model has 18M parameters and consists of a 1D convolutional layer and 3 LSTM layers.

\subsection{Training Description}

We utilized the Flower framework for both federated training and evaluation, given its user-friendly interface and active community \citep{Flower}. Additionally, we employed Hugging Face's transformers library \citep{wolf2019huggingface} to load pre-trained models, and PyTorch as the underlying differentiation framework \citep{paszke2019pytorch}. We conducted training for each sparsely and densely activated transformer model for 30 epochs with the AdamW optimizer and (1e-3, 1e-4) learning rates. Our experiment runs in about 15-20 hours for dense transformers, while it takes a day for MoE training. For FL, we assigned four clients and conducted five rounds of training, with each client training for one epoch per round for both the Non-IID and IID setups. All these experiments are carried out on 256 GB of RAM and two NVIDIA A40 GPUs, an Intel(R) Xeon(R) Gold 6338 machine.

\section{Experimental Results}

Our experimental design consists of three stages:

\begin{enumerate}[noitemsep,nolistsep]
    \item Train the models under test in a centralized setting 
on the task of emoji prediction. This setting is \textbf{Baseline}, which is done to ensure that the models under test are well-equipped for the task of emoji prediction before federated and centralized comparisons.
    
    \item Take the trained models and train them in FL with a new dataset distributed to clients. In this setup, we mainly carry out an experiment while assuming that all clients have clean data (clean scenario) and another experiment while assuming that some clients have data that has been label-flipped (attack scenario). In both scenarios, we carry out experiments for both \textbf{IID} and \textbf{Non-IID} FL to simulate real scenarios. For further clarification, see Figure \ref{fig:FL_abc}.
    \begin{figure}[htbp]
    \centering
    \includegraphics[width=0.5\textwidth]{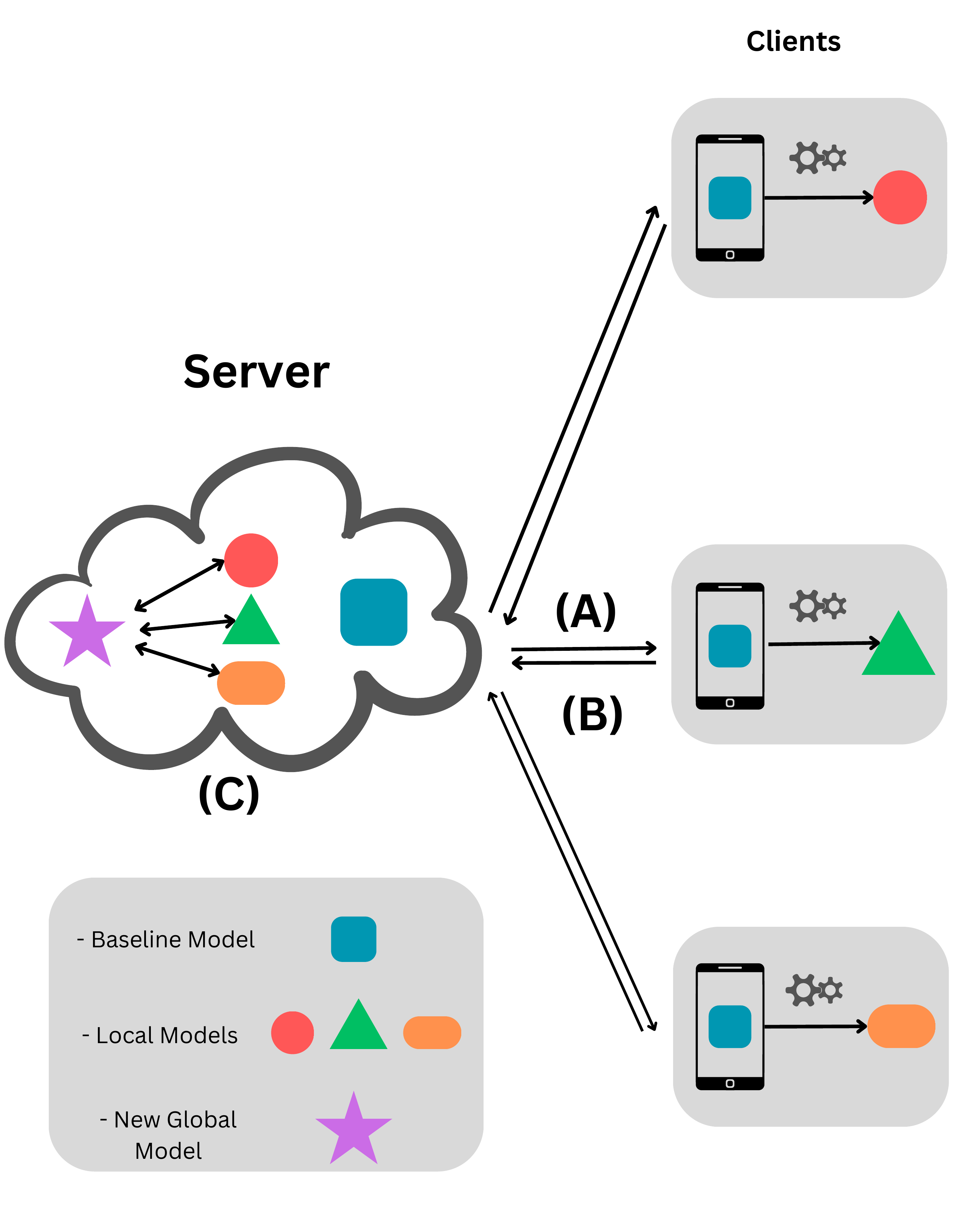}
    \caption{Second stage of our experimental design. (A) shows how the server pushes the baseline model to the clients for training on their local dataset. (B) demonstrates how the clients push back the tuned model to the server. Finally, (C) represents the process by which the server combines all models from the clients to build a new global model.}
    \label{fig:FL_abc}
    \end{figure}
    
    \item Take the trained models from the first stage and fine-tune these models in a centralized setting on the combination of all the distributed data used to train the FL. The third stage is the \textbf{Finetuned} setting, which is done to compare the performance of the federated learning distributed learning approach versus the traditional centralized learning approach in the task of emoji prediction. Our training set is divided into two halves. The first half of the data is used for the Baseline stage, while the second is used in either the FL stage (both IID and non-IID) or the Finetuned stage. We carry out all these experiments in unilingual and multilingual setups while measuring the corresponding Macro-F1 score for the 20-class emojis. 
\end{enumerate}

Macro-F1 and Micro-F1 are two evaluation metrics used in multi-class classification problems. The Micro-F1 score gives equal weight to each individual instance in the dataset, while the Macro-F1 score gives each class an equal weight. In cases where we have unbalanced classes, the Macro-F1 score is often used as it gives equal importance to each class, regardless of its frequency in the dataset.

\subsection{Experimenting with the Number of Clients in Federated Learning}

We present the results of our initial experiments to determine the optimal number of clients in federated learning for our problem. We used an LSTM IID multilingual model and evaluated its performance on: Micro-F1 and Macro-F1 scores.

\begin{table}[http]
\centering
\begin{tabular}{|c| c| c| c|}
\hline 
Metrics &  Baseline & IID & Finetuned
\\ \hline
Accuracy & 43.4\% & 45.1\% & 44.8\% \\
Macro-F1 & 27.6\% & 29.2\% & 29.1\% \\
\hline
\end{tabular}
\caption{Multilingual Centralized and Federated with 4 clients Accuracy and Macro-F1 Scores for SemEval test dataset.}
\label{tab:x_label}
\end{table}

\begin{figure}[htbp]
\centering
\includegraphics[width=0.45\textwidth]{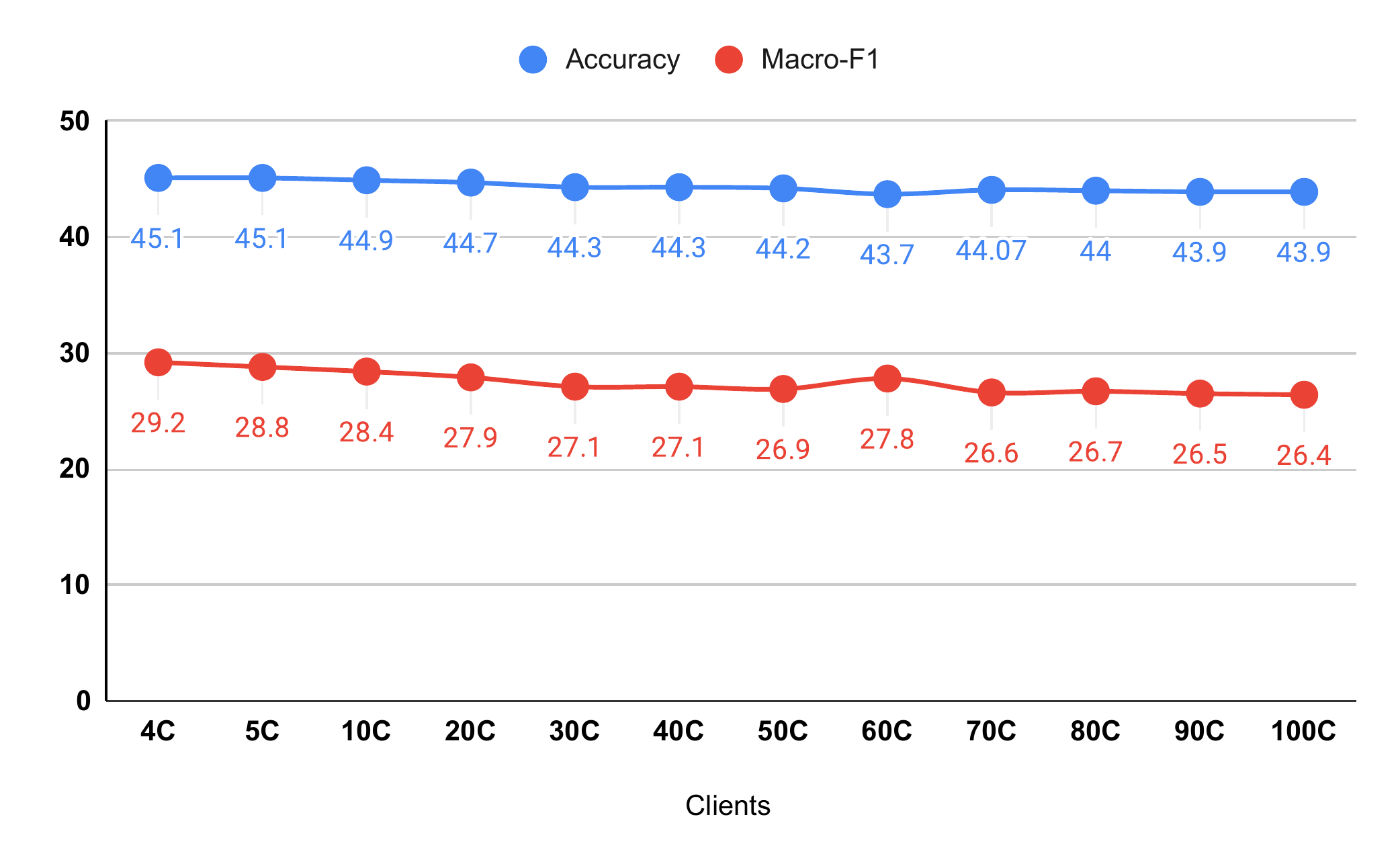}
\caption{SemEval Test Results of LSTM IID Multilingual experi-
ments with varying clients.}
\label{fig:mychart}
\end{figure}

As shown in Figure \ref{fig:mychart}, we observe that FL shows consistent performance across different number of clients in terms of Accuracy and Macro-F1 scores. There are slight drops in Macro-F1 because we use a fixed dataset distributed into multiple clients leading to a smaller data share per client. Smaller data may yield lower scores, which was also observed in the literature \citep{caldarola2022improving}.

So our initial experiments suggest that federated learning can improve the performance of our model, and a smaller number of clients might be more appropriate for our problem.

However, we can obtain higher scores by using transformer models, which is what we will do in the upcoming experiments.

\subsection{Unilingual and Multilingual Macro-F1 results in clean scenario}

Table \ref{tab:uni_clean_results} presents the results of the unilingual models for the SemEval English test dataset. As can be seen, The Finetuned models generally perform better than their baseline and IID counterparts. For example, the Bert-Base model has a baseline score of 36.9\%, while the IID score is 37.4\%, and the Finetuned score is 38.1\%. This shows that while the Finetuned process improves the model's performance, the IID training approach is also effective and can achieve a score close to the Finetuned model.

\begin{table}[http] \centering
\centering
{\small
\begin{tabular}{|c| c| c| c|}
\hline 
Model &  Baseline & IID & Finetuned
\\ \hline
Switch-Base-8 & 33.2\% & 37.3\% & 36.6\% \\
Bert-Base & 36.9\% & 37.4\% & 38.1\% \\
XLM-R & 35.9\% & 36.7\% & 37.6\% \\
M-MiniLM & 33.3\% & 33.9\% & 35.9\% \\
\hline
\end{tabular}
}
\caption{Unilingual Centralized and Federated Macro-F1 Accuracy Scores for SemEval test dataset.}
\label{tab:uni_clean_results}
\end{table}

Turning to \textit{multilingual}, Table \ref{tab:multi_clean_results} presents the Macro-F1 accuracy scores for centralized and FL multilingual models on the SemEval English dataset. The baseline accuracy for all models is improved by applying federated learning with the IID, which achieves slightly higher accuracy than the Non-IID setting. 
This suggests that data distribution has an impact on model performance. Additionally, finetuning the models further improves the Baseline performance with only small differences observed between federated and Finetuned results.

\begin{table}[http] \centering
{\small
\begin{tabular}{ | c|  c| c| c| c|}
\hline
Model &  Baseline & IID & Non-IID & Finetuned
\\ \hline
Bert-Base & 35.3\% & 36.8\% & 36.3\% & 35.7\% \\
XLM-R & 33.4\% & 34.9\% & 34.9\% & 34.1\% \\
M-MiniLM  & 31.4\% & 32.7\% & 32.03\% & 32.9\% \\
\hline
\end{tabular}
}
\caption{Multilingual Centralized and Federated Macro-F1 Accuracy Scores for SemEval English Dataset.}
\label{tab:multi_clean_results}
\end{table}

When comparing the unilingual and multilingual models via SemEval, we can see a drop in the Macro-F1 scores due to the mixed languages in the multilingual dataset. However, the drop was not significant, indicating that the models can handle multiple languages to a certain extent. Overall, the federated approach has shown to be a viable alternative to the centralized approach in terms of performance.

Using the Twitter multilingual dataset, Table \ref{tab:multi_clean_twitter_results} outlines the trained \textit{multilingual} Macro-F1 scores in centralized and FL setups. Similar observations to the case of SemEval are seen.

\begin{table}[http] \centering
{\small
\begin{tabular}{|c| c| c| c| c|}
\hline
Model     & \multicolumn{4}{|c|}{Average Test Results for Twitter Dataset} \\ \hline
 & Baseline & IID & Non-IID     & Finetuned   \\ \cline{2-5} 
Bert-Base & 29.2\%  & 30.6\%  & 30.7\%      & 30.2\% \\
XLM-R     & 26.5\%  & 28.2\%  & 28.02\% & 28.3\% \\
M-MiniLM  & 24.2\%  & 25.7\%  & 26.1\%      & 25.9\% \\ \hline
\end{tabular}
}
\caption{Centralized and Federated Learning average results for the Twitter Multilingual (e.g Spanish, French, Italian) Dataset.}
\label{tab:multi_clean_twitter_results}
\end{table}

In Table \ref{tab:multi_clean_twitter_bert_ber_lang_results}, we show the Macro-F1 score per language of the best-performing multilingual model Bert-Base in the centralized and federated setups. As shown, the accuracy does not significantly vary per language, which shows the effectiveness of our trained models.

\begin{table}[http] \centering
{\small
\begin{tabular}{|c|c|c|c|c|}
\hline
Data & Baseline & IID    & Non-IID & Finetuned \\ \hline
Spanish                      & 27.7\%    & 28.5\% & 28.8\%  & 27.3\%    \\ \hline
French                       & 29.6\%   & 31.3\% & 31.1\%  & 30.9\%    \\ \hline
Italian                      & 30.2\%   & 32.2\% & 32.3\%  & 32.4\%    \\ \hline
\end{tabular}
}
\caption{Centralized and Federated Learning Results for the Twitter multilingual dataset for Bert-Base model.}
\label{tab:multi_clean_twitter_bert_ber_lang_results}
\end{table}

\subsection{Comparison between our models and the literature}

Looking at the results in Table \ref{tab:other_papers_results}, we can observe that most of our multilingual models perform better than the majority of models that were also trained on the SemEval training dataset from the literature in terms of both Micro-F1 and Macro-F1. Specifically, our multilingual models in most cases achieved more than 36\% Macro-F1 or more than 49\% Micro-F1, whereas the best-performing model from the literature, BERT(Twitter) \citep{edwards2020go}, achieved 40\% Micro-F1. Furthermore, our unilingual model with Bert-Base achieved 38.1\% Macro-F1, which is comparable to the performance of BERT(Twitter), which achieved 38\% Macro-F1. Moreover, our federated models achieved more than 50\% Micro-F1, which is better than the performance of BERT(Twitter). Following the Large Language Models interest, we carried out an experiment using the Davinci-003 model \cite{brown2020language} on the SemEval set in zero-shot. Davinci-003 achieved a Macro-F1 score of 16\%, which also shows the promise of our trained FL models.

\begin{table}[http] \centering
{\small
\begin{tabular}{|c|c|c|c|}
\hline
Model                               & Micro-F1 & Macro-F1 \\ \hline
BiLSTM                              & 29.6\%  & 21.3\%  \\
\citep{venkit2021asourceful}        &   &   \\ \hline
Proposed LSTM IID                   & 45.1\%  & 29.2\%  \\ 
Multilingual                        &   &   \\ \hline
XLM-Tw                              & -        & 30.9\%  \\ 
\citep{barbieri2022xlm}             &         &   \\ \hline
TweetNLP                            & -        & 34.0\%  \\ 
\citep{camacho2022tweetnlp}         &        &   \\ \hline
SemEval first team                  & 47.1\%  & 35.9\%  \\ 
\citep{barbieri2018semeval}         &   &   \\ \hline
BERT (Twitter)                      & 40.0\%  & 38.0\%  \\ 
\citep{edwards2020go}               &   &   \\ \hline
Proposed Bert-Base Fintuned         & 49.4\%  & 35.7\%  \\ 
Multilingual                        &   &   \\ \hline
Proposed Bert-Base FL Non-IID       & 49.5\%  & 36.3\%  \\ 
Multilingual                        &   &   \\ \hline
Proposed Bert-Base FL IID           & 50.3\%  & 36.8\%  \\ 
Multilingual                        &   &   \\ \hline
Proposed Bert-Base FL IID           & 50.9\%  & 37.4\%  \\ 
Unilingual                          &   &   \\ \hline
Proposed Bert-Base Fintuned         & 50.1\%  & 38.1\%  \\ 
Unilingual                          &   &   \\ \hline

\end{tabular}
}
\caption{Comparison between our models' performance and models from the literature on emoji prediction task using the Micro-F1 and Macro-F1 metrics.}
\label{tab:other_papers_results}
\end{table}

\subsection{Multilingual Macro-F1 Results on unseen language}

To investigate the trained models' performance on unseen languages (i.e, zero-shot), we run inference on an unseen German dataset in Baseline, FL, and Finetune settings. Table \ref{tab:multi_clean_unseen_results} shows that there is some drop in performance due to the zero-shot setting. However, this experiment still shows that FL performs at least similarly to centralized settings even in unseen languages.

\begin{table}[http] \centering
{\small
\begin{tabular}{| c | c| c| c| c|}
\hline
Model & Baseline & IID & Non-IID & Finetuned \\ \hline
Bert-Base & 21.9\% & 23.1\% & 23.1\% & 21.5\% \\
XLM-R & 20.04\% & 20.9\% & 21.07\% & 19.2\% \\
M-MiniLM & 15.4\% & 16.5\% & 16.4\% & 17.1\% \\ 
\hline
\end{tabular}
}
\caption{The zero-shot inference results for Centralized and Federated Learning }
\label{tab:multi_clean_unseen_results}
\end{table}

\subsection{Multilingual Macro-F1 Results in Label-flipping Attack Scenario}

Table \ref{tab:Multilingual_50_attack_ratio} and \ref{tab:Multilingual_50_attack_ratio_it_fr_sp_avg} depict the results of label flipping experiments when 50\% clients are attacked (i.e 50\% of the data is attacked) using the SemEval and Twitter datasets, respectively. The tables compare the centralized and FL results for FedAVG and Krum. The results demonstrate that Krum performed better than FedAVG in both datasets, with higher accuracy rates. Krum was able to handle the label-flipping attack scenario and produced scores that were very close to the results obtained with the Finetune setting. 

\begin{figure}
	\centering
            \begin{tikzpicture}[scale=0.8,]
            \begin{axis}[
            	x tick label style={},
            	ylabel=Macro-F1,
            	enlargelimits=0.05,
            	legend style={at={(0.5,-0.1)},
            	anchor=north,legend columns=-1},
            	ybar interval=0.7,
                xtick={1,2,3},
                xticklabels={Fed-IID,Fed-Non-IID,C},
                title={Bert-Base Model},
            ]
            \addplot 
            	coordinates {(1,26.2) (2,28.1)
            		 (3,18.5) };
            \addplot 
            	coordinates {(1,36.5) (2,35.2) 
            		(3,18.2) };
            
            \tikzset{myline/.style={black, line width=1pt}}
            \draw[myline] (axis cs:1,24.4) -- node[above] {Finetuned} (axis cs:3,24.4);
            
            \legend{FedAVG,Krum}
            \end{axis}
            \end{tikzpicture}
    \centering
	\caption{Centralized and Federated Learning Results in Label-Flipping Attack Scenario for the SemEval English test dataset for Bert-Base Model.}
\label{tab:Multilingual_50_attack_ratio_Bert_Figure}
\end{figure}
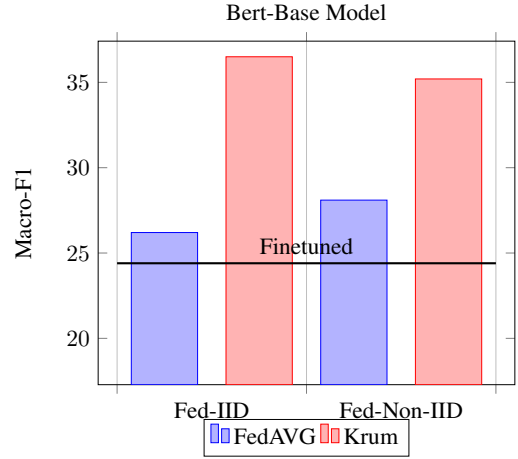

Table \ref{tab:Multilingual_50_attack_ratio} presents the experiment results for three different models, namely Bert-Base, XLM-R, and M-MiniLM. However, we will focus on the results of the Bert-Base model in Figure \ref{tab:Multilingual_50_attack_ratio_Bert_Figure}, where Bert achieved 36.8\% Macro-F1 in the clean IID scenario but dropped to 26.2\% under FedAVG with Fed-IID due to label-flipping attacks. Traditional training and FL with FedAVG had low Macro-F1 of 24.4\%, while Krum aggregation function achieved 36.5\% Macro-F1, showing superior handling of label-flipping attacks and improving FL model performance. Appendix \ref{app:Multilingual_25_attack_ratio} shows similar results for 25\% Toxic clients experiments.

\begin{table}[http] \centering
{\small
\begin{tabular}{|c | c | c | c |}
\hline
 Model            & Setting     & \multicolumn{1}{c|} {FedAVG} & Krum    \\ \hline
                  & Fed-IID     & \multicolumn{1}{c|} {26.2\%} & 36.5\%  \\
 Bert-Base        & Fed-Non-IID & \multicolumn{1}{c|} {28.1\%} & 35.2\%  \\ \cline{3-4}
                  & Finetuned   & \multicolumn{2}{c|} {24.4\%}           \\ \hline
                  &             & \multicolumn{1}{c|} {FedAVG} & Krum    \\ \cline{3-4}
                  & Fed-IID     & \multicolumn{1}{c|} {27.6\%} & 34.8\%  \\
 XLM-R            & Fed-Non-IID & \multicolumn{1}{c|} {27.5\%} & 32.9\%  \\ \cline{3-4}
                  & Finetuned   & \multicolumn{2}{c|} {23.3\%}           \\ \hline
                  &             & \multicolumn{1}{c|} {FedAVG} & Krum    \\ \cline{3-4}
                  & Fed-IID     & \multicolumn{1}{c|} {25.3\%} & 32.6\%  \\
 M-MiniLM         & Fed-Non-IID & \multicolumn{1}{c|} {26.7\%} & 30.5\%  \\ \cline{3-4}
                  & Finetuned   & \multicolumn{2}{c|} {23.9\%}           \\ \hline

\end{tabular}
}
\caption{Centralized and Federated Learning Results in Label-Flipping Attack Scenario for the SemEval English test dataset.}
\label{tab:Multilingual_50_attack_ratio}
\end{table}

\begin{table}[http] \centering
{\small
\begin{tabular}{| c | c | c | c |}
\hline
 Model            & Setting     & \multicolumn{1}{c|} {FedAVG} & Krum    \\ \hline
                  & Fed-IID     & \multicolumn{1}{c|} {21.8\%} & 30.1\%  \\
 Bert-Base        & Fed-Non-IID & \multicolumn{1}{c|} {23.6\%} & 29.4\%  \\ \cline{3-4}
                  & Finetuned   & \multicolumn{2}{c|} {21.8\%}           \\ \hline
                  &             & \multicolumn{1}{c|} {FedAVG} & Krum    \\ \cline{3-4}
                  & Fed-IID     & \multicolumn{1}{c|} {20.4\%} & 27.3\%  \\
 XLM-R            & Fed-Non-IID & \multicolumn{1}{c|} {21.5\%} & 26.6\%  \\ \cline{3-4}
                  & Finetuned   & \multicolumn{2}{c|} {20.8\%}           \\ \hline
                  &             & \multicolumn{1}{c|} {FedAVG} & Krum    \\ \cline{3-4}
                  & Fed-IID     & \multicolumn{1}{c|} {19.02\%} & 24.9\%  \\
 M-MiniLM         & Fed-Non-IID & \multicolumn{1}{c|} {20.9\%} & 23.9\%  \\ \cline{3-4}
                  & Finetuned   & \multicolumn{2}{c|} {20.3\%}           \\ \hline

\end{tabular}
}
\caption{Centralized and Federated Learning Results in Label-Flipping Attack Scenario for the average results for the Twitter multilingual dataset.}
\label{tab:Multilingual_50_attack_ratio_it_fr_sp_avg}
\end{table}

Similar to the experiment in Table \ref{tab:Multilingual_50_attack_ratio}. Table \ref{tab:Multilingual_50_attack_ratio_it_fr_sp_avg} shows that Krum outperforms FedAVG in both the Non-IID and IID settings as well. Also in Table \ref{tab:multi_clean_german_results} reports the results of applying label flipping to the German Zero-shot scenario,

\begin{table}[http] \centering
{\small
\begin{tabular}{| c | c | c | c |}
\hline
 Model            & Setting     & \multicolumn{1}{c|} {FedAVG} & Krum    \\ \hline
                  & Fed-IID     & \multicolumn{1}{c|} {15.3\%} & 23.5\%  \\
 Bert-Base        & Fed-Non-IID & \multicolumn{1}{c|} {15.7\%} & 21.3\%  \\ \cline{3-4}
                  & Finetuned   & \multicolumn{2}{c|} {16.3\%}           \\ \hline
                  &             & \multicolumn{1}{c|} {FedAVG} & Krum    \\ \cline{3-4}
                  & Fed-IID     & \multicolumn{1}{c|} {14.1\%} & 20.9\%  \\
 XLM-R            & Fed-Non-IID & \multicolumn{1}{c|} {14.04\%} & 19.8\%  \\ \cline{3-4}
                  & Finetuned   & \multicolumn{2}{c|} {14.3\%}           \\ \hline
                  &             & \multicolumn{1}{c|} {FedAVG} & Krum    \\ \cline{3-4}
                  & Fed-IID     & \multicolumn{1}{c|} {11.3\%} & 16.2\%  \\
 M-MiniLM         & Fed-Non-IID & \multicolumn{1}{c|} {11.9\%} & 14.3\%  \\ \cline{3-4}
                  & Finetuned   & \multicolumn{2}{c|} {11.7\%}   \\ \hline

\end{tabular}
}
\caption{Centralized and Federated Learning Results in Label-Flipping Attack Scenario for the German Zero-shot.}
\label{tab:multi_clean_german_results}
\end{table}

\subsection{Comparing Results of Clean and Attack Scenarios using Various Aggregation Functions.}

Table \ref{tab:Bert_Base_Fed_only} provides valuable insights into the performance of a Bert-Base model in a federated multilingual setting under both clean and attack scenarios. The results highlight the impact of toxic clients on the federated setting, showing a decrease in performance for FedAVG, Fed-IID, and Fed-Non-IID. This decrease in performance is particularly evident when the toxic data constitutes 50\% of the overall data. However, the use of the Krum aggregation function can mitigate this drop in performance, as previously observed in literature.

Overall, this table serves to provide a clear comparison of the performance under different scenarios, which can be useful for identifying the most suitable aggregation function for a given federated learning scenario.

\begin{table}[http]
\centering
{\small
\begin{tabular}{| c c | c c |}
\hline
\multicolumn{2}{|c|}{{ Setting}} &
  \multicolumn{2}{c|}{Results} \\ \hline
\multicolumn{1}{|c|}{Cleaned} &
  Fed-IID & \multicolumn{2}{c|}{36.8} \\  
\multicolumn{1}{|c|}{FedAVG} &
  Fed-Non-IID & \multicolumn{2}{c|}{36.3} \\ \hline
\multicolumn{1}{|l|}{} &
  \multicolumn{1}{l|}{} &
  \multicolumn{1}{c|}{25\% Toxic} &
  50\% Toxic \\ \cline{3-4} 
\multicolumn{1}{|c|}{FedAVG} &
  Fed-IID &
  \multicolumn{1}{c|}{36.3} & 26.2 \\
\multicolumn{1}{|l|}{} &
  Fed-Non-IID &
  \multicolumn{1}{c|}{33.5} & 28.1 \\ \hline
\multicolumn{1}{|l|}{} &
  \multicolumn{1}{l|}{} &
  \multicolumn{1}{c|}{25\% Toxic} &
  50\% Toxic \\ \cline{3-4} 
\multicolumn{1}{|c|}{Krum} &
  Fed-IID &  \multicolumn{1}{c|}{36.5} & 36.5 \\
\multicolumn{1}{|l|}{} &
  Fed-Non-IID &  \multicolumn{1}{c|}{35.3} & 35.2 \\ \hline
\end{tabular}
}
\caption{Macro-F1 Results of Bert-Base Model in Federated Multilingual Setting under Clean and Attack Scenarios for the SemEval English test dataset.}
\label{tab:Bert_Base_Fed_only}
\end{table}

\subsection{Computational Overhead Comparison between Krum and FedAVG}




The exact computational overhead of the Krum algorithm compared to FedAvg or other aggregation algorithms depends on several factors, such as the number of participating clients, the size of the models, and the specific implementation of the algorithms. In some cases, the computational overhead of the Krum algorithm may be similar to that of FedAvg or even lower, depending on the specific scenario. Table\ref{tab:Resource_constraints} shows that the Fed-Non-IID algorithm takes longer than the Fed-IID algorithm due to the varying sizes of language datasets among clients. The Krum algorithm takes $\sim$10\% more time than FedAVG in our experiments.

\begin{table}[http] \centering
{\small
\begin{tabular}{| c | c | c | c |}
\hline
\multicolumn{2}{|c|}{ Setting}  & Time (Hours)    \\ \hline
\multirow{2}{*} {FedAVG}         & Fed-IID     & $\sim$18  \\
                                 & Fed-Non-IID & $\sim$22  \\ \hline
\multirow{2}{*} {Krum}           & Fed-IID     & $\sim$20  \\
                                 & Fed-Non-IID & $\sim$24  \\ \hline

\end{tabular}
}
\caption{Comparison of computational overhead between FedAVG and Krum algorithms in terms of training time for each federated learning experiment.}
\label{tab:Resource_constraints}
\end{table}



To estimate the communicated payload for each client in our federated learning model, the total size of model parameters needs to be calculated. Assuming that the local model is the M-MiniLM model of size 0.47 GB, and there are 4 clients participating in each round, then each client needs to transmit 0.47 GB of data to the server during each round of training. Since there are 5 rounds in total, the total amount of data transmitted per client for one epoch would be 2.35 GB (0.47 GB x 5). This is an approximate estimate and does not take into account factors such as compression techniques or network latency. The actual resource constraints may vary depending on these factors.

Table \ref{tab:data_transmitted} shows the estimated amount of data transmitted during one epoch of federated learning using different models. For the M-MiniLM model, the estimated amount of data transmitted per client for one epoch would be 2.35 GB, which is within the resource constraints for modern devices.

\begin{table}[http] \centering
{\small
\begin{tabular}{|c|c|c|}
\hline
{ Model Name} &
{ Model Size} &
{ Data Transmitted} \\ \hline
{ Switch-Base-8} & { 1.24 GB}  & { 1.24*5 = 6.2 GB}  \\
{ Bert-Base}     & { 1.12 GB}  & { 1.12*5 = 5.6 GB}  \\
{ XLM-R}         & { 1.11 GB}  & { 1.11*5 = 5.55 GB}  \\
{ M-MiniLM}      & { 0.47 GB} & { 0.47*5 = 2.35 GB} \\ \hline
\end{tabular}
}
\caption{Estimated amount of data transmitted per client during one round of FL.}
\label{tab:data_transmitted}
\end{table}

\section{Conclusion}
\label{sec:Conclusion}

This paper proposes federated learning-based multilingual emoji prediction in clean and attack scenarios. Different transformer models with varying sizes are trained in centralized and federated for which we compare their corresponding emoji prediction accuracy. Our experiments were carried out in seen and unseen languages using different data sources and distributions. Due to federated performance, federated learning can act as a substitute for centralized settings to gain privacy and access to multiple data sources benefits. In addition, we showed that our federated learning performance is competitive with the SemEval shared task on multilingual emoji prediction. In the future, we wish to explore how to achieve similar accuracy performance while considering the communication efficiency of federated learning \citep{passban2022training}. We also believe that more active research in federated learning user personalization \citep{arivazhagan2019federated} given the subjectivity of emojis can be investigated.

\section*{Acknowledgements}
We sincerely thank our invaluable contributors for their unwavering support during the entire project. First and foremost, we thank Dr. Mohamed Afify, Principal Applied Scientist at Microsoft Advanced Technology Lab, for his continuous guidance and encouragement. Dr. Mona Farouk for her diligent supervision and contributions to the DEBI program. We extend our thanks to Dr. Yuanzhu Chen, Professor at Queen's University, for generously providing us with his workstation. In addition, we would like to thank Dr. Muhammad Jabreel for his valuable suggestions on attack scenarios. Moreover, we express our appreciation to Abdelrahman ElHamoly for his invaluable assistance in the initial phase. Finally, we would like to extend our thanks to Orion Weller for his assistance and prompt responses to our inquiries about federated learning.

\bibliography{anthology,custom}
\bibliographystyle{acl_natbib}

\appendix
\section{Multilingual Macro-F1 Results in Label-flipping Attack Scenario Results}
\label{app:Multilingual_25_attack_ratio}

Tables \ref{tab:SemEval_label_flipping_25} and \ref{tab:twitter_label_flipping_25} present a comparison of centralized and federated learning for different multilingual models in a label-flipping attack scenarios with an Attack Ratio of 25\%. The models' performance was evaluated using two different settings, Fed-IID and Fed-Non-IID, and two federated learning algorithms, FedAVG and Krum. Both tables show that the models' performance decreases in the label-flipping attack scenario, and is worse for the Fed-Non-IID setting than for the Fed-IID setting. Furthermore, Krum outperforms the FedAVG aggregator in all cases, and the Bert-Base model generally performs better than the other models.

\begin{table}[http] \centering
{\small
\begin{tabular}{|c | c | c  c |}
\hline
 Model            & Setting     & \multicolumn{1}{c|} {FedAVG} & Krum    \\ \hline
                  & Fed-IID     & \multicolumn{1}{c|} {36.3\%} & 36.5\%  \\
 Bert-Base        & Fed-Non-IID & \multicolumn{1}{c|} {33.5\%} & 35.3\%  \\ \cline{3-4}
                  & Finetuned   & \multicolumn{2}{c|} {35.6\%}           \\ \hline
                  &             & \multicolumn{1}{c|} {FedAVG} & Krum    \\ \cline{3-4}
                  & Fed-IID     & \multicolumn{1}{c|} {34.7\%} & 34.9\%  \\
 XLM-R            & Fed-Non-IID & \multicolumn{1}{c|} {32.7\%} & 33.4\%  \\ \cline{3-4}
                  & Finetuned   & \multicolumn{2}{c|} {33.9\%}           \\ \hline
                  &             & \multicolumn{1}{c|} {FedAVG} & Krum    \\ \cline{3-4}
                  & Fed-IID     & \multicolumn{1}{c|} {32.3\%} & 32.2\%  \\
 M-MiniLM         & Fed-Non-IID & \multicolumn{1}{c|} {30.4\%} & 30.7\%  \\ \cline{3-4}
                  & Finetuned   & \multicolumn{2}{c|} {32.7\%}           \\ \hline

\end{tabular}
}
\caption{Centralized and Federated Learning Results in Label-Flipping Attack Scenario for the SemEval English test dataset.}
\label{tab:Multilingual_25_attack_ratio}
\label{tab:SemEval_label_flipping_25}
\end{table}

\begin{table}[http] \centering
{\small
\begin{tabular}{| c | c | c | c |}
\hline
 Model            & Setting     & \multicolumn{1}{c|} {FedAVG} & Krum    \\ \hline
                  & Fed-IID     & \multicolumn{1}{c|} {28.2\%} & 29.1\%  \\
 Bert-Base        & Fed-Non-IID & \multicolumn{1}{c|} {30.5\%} & 29.7\%  \\ \cline{3-4}
                  & Finetuned   & \multicolumn{2}{c|} {28.9\%}           \\ \hline
                  &             & \multicolumn{1}{c|} {FedAVG} & Krum    \\ \cline{3-4}
                  & Fed-IID     & \multicolumn{1}{c|} {26.1\%} & 27.7\%  \\
 XLM-R            & Fed-Non-IID & \multicolumn{1}{c|} {27.1\%} & 27.3\%  \\ \cline{3-4}
                  & Finetuned   & \multicolumn{2}{c|} {26.7\%}           \\ \hline
                  &             & \multicolumn{1}{c|} {FedAVG} & Krum    \\ \cline{3-4}
                  & Fed-IID     & \multicolumn{1}{c|} {22.9\%} & 25.3\%  \\
 M-MiniLM         & Fed-Non-IID & \multicolumn{1}{c|} {24.8\%} & 24.6\%  \\ \cline{3-4}
                  & Finetuned   & \multicolumn{2}{c|} {24.6\%}           \\ \hline

\end{tabular}
}
\caption{Centralized and Federated Learning Results in Label-Flipping Attack Scenario for the average results for the Twitter multilingual dataset.}
\label{tab:twitter_label_flipping_25}
\end{table}

Table \ref{tab:German_25_attack} presents a comparison between centralized and federated learning approaches using FedAVG and Krum aggregation functions in a label-flipping attack scenario for three different models. We observed that Krum outperforms FedAVG in all cases, indicating that the choice of aggregation function has a significant impact on the model's performance. However, we also found that the model's performance is heavily dependent on the architecture and the type of aggregation function used.

The results of our experiments highlight the vulnerability of federated learning to label-flipping attacks. This vulnerability emphasizes the importance of carefully selecting the federated learning algorithm and aggregation function to mitigate such attacks. Additionally, our experiments revealed that the Fed-IID setting is less vulnerable to label-flipping attacks, which suggests that data distribution plays a critical role in the performance of federated learning models.

To further investigate the impact of label-flipping attacks on federated learning models, future experiments can explore the impact of different attack ratios and other attack scenarios. Moreover, it would be interesting to study the impact of other factors, such as the heterogeneity of data sources and the distribution of data samples, on the vulnerability of federated learning models to attacks. By gaining a better understanding of the vulnerabilities of federated learning, we can develop more robust and secure models that are better suited for real-world applications.

\begin{table}[http]
\centering
{\small
\begin{tabular}{| c | c |c | c|}
\hline
 Model            & Setting     & \multicolumn{1}{c|} {FedAVG} & Krum    \\ \hline
                  & Fed-IID     & \multicolumn{1}{c|} {20.67\%} & 23.48\%  \\
 Bert-Base        & Fed-Non-IID & \multicolumn{1}{c|} {20.60\%} & 21.62\%  \\ \cline{3-4}
                  & Finetuned   & \multicolumn{2}{c|} {21.6\%}           \\ \hline
                  &             & \multicolumn{1}{c|} {FedAVG} & Krum    \\ \cline{3-4}
                  & Fed-IID     & \multicolumn{1}{c|} {18.60\% } & 21.25\%  \\
 XLM-R            & Fed-Non-IID & \multicolumn{1}{c|} {18.81\%} & 19.93\%  \\ \cline{3-4}
                  & Finetuned   & \multicolumn{2}{c|} {18.2\%}           \\ \hline
                  &             & \multicolumn{1}{c|} {FedAVG} & Krum    \\ \cline{3-4}
                  & Fed-IID     & \multicolumn{1}{c|} {13.78\%} & 16.47\%  \\
 M-MiniLM         & Fed-Non-IID & \multicolumn{1}{c|} {14.62\%} & 15.38\%  \\ \cline{3-4}
                  & Finetuned   & \multicolumn{2}{c|} {14.7\%}           \\ \hline

\end{tabular}
}
\caption{Centralized and Federated Learning Results in Label-Flipping Attack Scenario for the German Zero-shot.}
\label{tab:German_25_attack}
\end{table}

\end{document}